\documentclass[11pt,a4paper]{article}
\usepackage{authblk}
\usepackage{emnlp2020}
\usepackage[utf8]{inputenc}
\DeclareUnicodeCharacter{0474}{}
\DeclareUnicodeCharacter{0405}{}
\DeclareUnicodeCharacter{041E}{}
\DeclareUnicodeCharacter{0408}{}
\usepackage{times}
\usepackage{latexsym}

\usepackage{multirow}
\usepackage{tabularx}
\usepackage{graphicx}
\usepackage{booktabs , amssymb}

\usepackage{subcaption}

\usepackage{microtype}

\aclfinalcopy % Uncomment this line for the final submission
\title{Type B Reflexivization as an Unambiguous Testbed\\for Multilingual Multi-Task Gender Bias}

\author[1]{\textbf{Ana Valeria Gonz{\'a}lez}}
\author[1,2]{\textbf{Maria Barrett}}
\author[2]{\textbf{Rasmus Hvingelby}}
\author[3]{\textbf{Kellie Webster}}
\author[1,3]{\\\textbf{Anders S{\o}gaard}}

\affil[ ]{$^1$University of Copenhagen, $^2$Alexandra Instituttet, $^3$Google Research}
\affil[ ]{\texttt{\{ana,mjb,soegaard\}@di.ku.dk},}
\affil[ ]{\texttt{rasmus.hvingelby@alexandra.dk}}
\affil[ ]{\texttt{websterk@google.com}}

\date{}

\begin{document}
\maketitle
\begin{abstract}
The one-sided focus on English in previous studies of gender bias in NLP misses out on opportunities in other languages: English challenge datasets such as GAP and WinoGender highlight model preferences that are "hallucinatory", e.g., disambiguating gender-ambiguous occurrences of 'doctor' as male doctors. We show that for languages with type B reflexivization, e.g., Swedish and Russian, we can construct multi-task challenge datasets for detecting gender bias that lead to {\em unambiguously wrong} model predictions: In these languages, the direct translation of 
'the doctor removed his mask' is not ambiguous between a coreferential reading and a disjoint reading. 
 Instead, the coreferential reading requires a non-gendered pronoun, and the gendered, possessive pronouns are {\em anti-reflexive}. We present a multilingual, multi-task challenge dataset, which spans four languages and four NLP tasks and focuses only on this phenomenon. We find evidence for gender bias across {\em all} task-language combinations 
 and correlate model bias with national labor market statistics.

\end{abstract}

\section{Introduction}

A reflexive pronoun is an anaphor that requires a $c$-commanding antecedent within its binding domain \cite{Chomsky:91}.\footnote{This means that the antecedent should be in the same sentence, be different from the pronoun and not command it, but any ancestor of the antecedent is an ancestor of the pronoun. This is why in {\em Lea$_1$'s sister$_2$ taught herself$_{1*/2/3*}$} the pronoun refers to {\em sister}, not to {\em Lea} or a discourse referent.} In languages with {\em Type B} reflexivization \cite{Heine:05}, the referent of a reflexive possessive pronoun has to be the subject of the clause, while non-reflexive possessive pronouns (so-called {\em anti-reflexives}) trigger an interpretation where its referent is {\em not}~the subject; see Table~\ref{fig:AB}.

\begin{table}
    \centering
    \begin{tabular}{l|ccc|ccc}
    \toprule &\multicolumn{3}{c}{\sc Type A}&\multicolumn{3}{c}{\sc Type B}\\
    \midrule
    Person&1st&2nd&3rd&    1st&2nd&3rd\\
    \midrule
    {\sc Refl}&\checkmark&\checkmark&\checkmark&\checkmark&\checkmark&\\
    \bottomrule
    \end{tabular}
    \caption{In Type B reflexivization \cite{Heine:05}, 3rd person pronouns cannot be used reflexively. We are interested in Type B languages with gendered pronouns, and where the non-gendered special (3rd person) reflexive marker has a possessive form.}
    \label{fig:AB}

\end{table}

We focus on the subset of those languages in which {\em anti-reflexive possessive pronouns are gendered, but reflexives are not}. This includes Chinese, Russian, Danish, and Swedish, as well as other Scandinavian, Slavic, and Sino-Tibetan languages languages \cite{Bily:81,Battistella:84,Kiparsky:01}.\footnote{This rules out languages such as German and French, where the reflexive (e.g., {\it sich} and {\it se}) does not have a possessive form \cite{Steinbach:98}. We focus on the reflexive and anti-reflexive {\it possessive} forms rather than pure reflexives, since they occur more freely, i.e., not only in the context of reflexive verbs, and they are thus more likely to interact with implicit gender assumptions.} Our motivation for highlighting this particular linguistic phenomenon is that the antecedents of reflexive and anti-reflexive pronouns are grammatically determined; if gender bias leads models (or humans) to predict alternative coreference chains, this violates hard grammatical rules and is thus a clear case of gender bias leading not only to 'hallucinations',\footnote{We use the word {\em hallucination} to refer to gender bias leading models to infer gender without evidence; see \newcite{Tian:ea:20} for a similar use of the term in abstractive summarization.} but to errors. 
To see this, consider the following examples: 

{\small
\begin{itemize}
    \item[(1)] The surgeon$_1$ put a book on {\sc Pron.Poss.Refl.3rd}$_1$ table. $\rightarrow$ The book is on the surgeon's$_1$ table.
    \item[(2)] The surgeon$_1$ put a book on {\sc Pron.Poss.3rd}$_2$ table. $\not\rightarrow$ The book is on the surgeon's$_1$ table.
\end{itemize}
}

Examples (1) and (2) should not be thought of as examples of English, but placeholders for sentences in the languages above since this grammatical distinction is not possible in English: the possessive reflexive ({\sc Pron.Poss.Refl.3rd}) and the possessive anti-reflexive ({\sc Pron.Poss.3rd}) in these languages would be translated to the same pronoun in English. In Example (1), the reflexive possessive pronoun is co-referential with the grammatical subject (as indicated by subscripts), which leads to the conclusion that the book is now on a table that is associated with the subject, in other words, the {\it surgeon's} table. In Example (2), in contrast, when an anti-reflexive possessive pronoun is used, this reading is no longer possible. Instead, Example (2) unambiguously means that the book is on someone else's table. This distinction is not possible in English where the same pronoun ({\em his}/{\em her}) would be used in both Examples (1) and (2): {\em The surgeon put a book on his table}, which is therefore ambiguous between a disjoint and a coreferent reading.

Language users may be {\em more likely to prefer the ungrammatical reflexive reading if the gender of the anti-reflexive possessive pronoun matches their (possibly gender-stereotypical) expectations about the referent of the subject}, in this case, {\em the surgeon}. A masculine possessive pronoun aligns with a prevalent stereotype that surgeons are men; although in the US, in reality, only 62\% are.\footnote{\url{http://www.bls.gov/cps/cpsaat11.htm}} 
Such a reading is, however, clearly not intended, and this is an example of when gender bias prohibits effective communication. Introducing a new referent in a discourse, usually comes at a cognitive cost when processing the sentence if the referent is not already salient \cite{Grosz:ea:95}. While Example (2) is grammatically unambiguous, language users may occasionally be willing to violate grammatical constraints to avoid the more costly non-coreferential reading, if the meaning of the grammatically correct disjoint reading does not align with their expectations about the world.\footnote{Note that this is {\em not} a conflict between syntax and semantics, such as, for example, those studied in \newcite{Kos:ea:10}, but a conflict between syntax, on the one hand, and belief bias and pragmatics.} 

The challenge dataset that we present here consists of examples such as the one above and is intended as a diagnostic of implicit gender assumptions in NLP models. It is applicable across four languages (Danish, Russian, Swedish, and Chinese) and four NLP tasks: natural language inference (NLI), machine translation (MT), coreference resolution, and language modeling (LM)). We will, for example, be interested in whether models are more likely to produce errors when the anti-reflexive pronouns -- {\sc Pron.Poss.3rd} in Example~(2) -- exhibit the gender that is implicitly associated with the entity in the subject position, i.e., {\em surgeon}. As should be clear by now, the challenge dataset is fundamentally different from previously introduced challenge datasets in that it focuses on a single linguistic phenomenon that exists across many languages \cite{lodrup2011norwegian,honselaar1986reflections,cohen1973hindi,stoykova2012inflectional} and includes {\bf four languages and four tasks}, and because it focuses on gender bias leading to prediction errors rather than 'hallucinations', i.e., unwarranted disambiguations. To the best of our knowledge, the dataset introduced below is in this way the first of its kind. 

\paragraph{Contributions} We present a multilingual, multi-task challenge dataset focusing on a specific linguistic phenomenon found in some Scandinavian, Slavic, and Sino-Tibetan languages, namely {\it gendered possessive anti-reflexive pronouns} in combination with non-gendered possessive reflexive pronouns. We show, by designing multilingual example generation templates by hand, how this phenomenon can interact with gender assumptions in interesting ways. This results in a unique challenge dataset, which we use to detect and quantify gender biases in state-of-the-art and off-the-shelf models across several tasks, including machine translation, natural language inference, coreference resolution, and language modeling. Unlike all other previous challenge datasets focusing on gender bias, our examples quantify {\em to what extent gender bias in models leads to prediction errors}, rather than unwarranted disambiguation. Data and code is available at \url{https://github.com/anavaleriagonzalez/ABC-dataset}

\section{The Anti-reflexive Bias Challenge}

The {\sc Anti-reflexive Bias Challenge} (ABC) dataset is designed to force humans and models to align with either widespread gender assumptions or hard grammatical rules. Note, again, that this is in sharp contrast with other gender bias challenge datasets, where gender biases lead to biases in semantic disambiguation, but do not interact with grammatical constraints. Our approach is similar to previous work in other respects: 

Similarly to \newcite{rudinger2018gender} and other recent challenge datasets, ABC relies on hand-written templates, which are used to generate examples in conjunction with lists of occupations. 
We make use of the 60 occupations listed in \newcite{caliskan2017semantics} containing statistics about gender distributions across professions, taken from the U.S. Bureau of Labor Statistics. 
Specifically, we generate a base set of 4,560 sentences from 38 templates, two tenses (present and past), and 60 occupations. The 38 templates vary the position of the pronouns, e.g.: 

{\small
\begin{itemize}
    \item[(3)] The {\sc Occupation} lost {\sc Pron.Poss.3rd} wallet at the house.
    \item[(4)] {The {\sc Occupation} lost the wallet at {\sc Pron.Poss.3rd} house}.
\end{itemize}}

where {\sc Pron.Poss.3rd}, in this case, is a place holder for anti-reflexive and reflexive third-person pronouns. Our templates only include transitive verbs. 

In our language modelling experiments, we predict the pronoun in question. For NLI and coreference, we introduce three variations of each datapoint (possessive masculine, possessive feminine (anti-reflexive) pronouns and the non-gendered reflexive pronoun). This leads to a total of 13,680 examples for each language. For NLI, we use these as premises and add possible entailments to our templates. See Examples (1) and (2). For machine translation, we use the English versions of Examples (3) and (4) as source sentences, with feminine and masculine third-person pronouns. This leads to 9,120 translation problems. Native speakers manually verified and corrected all templates and sample examples for all tasks. Appendix A shows examples from the four tasks in the four languages. We discuss each task in detail below. 
 
\paragraph{NLI} Examples (1) and (2) illustrate the entailment phenomenon that we are interested in. Reflexive possessive pronouns are coreferential with their subjects, which leads to the interpretation that the book is on the surgeon's table. Anti-reflexive pronouns, on the other hand, prevent this reading and leads to an interpretation that a new discourse entity -- another person -- exists and that the book is located on that person's table. 

The general form of our inference examples is as follows: 

{\small
\begin{itemize}
    \item[(5)] {\sc Occupation.DEF}$_1$ [{\sc Verb Phrase}] {\sc Pron.Poss.Refl.3rd}$_1$ {\sc Object} {\sc Prep} {\sc Noun.DEF}. $\rightarrow$ {\sc Occupation.DEF.POSS}$_1$ {\sc Object} [{\sc Verb Phrase.PASSIVE}] {\sc Prep} {\sc Noun.DEF}.
    \item[(6)] {\sc Occupation.DEF}$_1$ [{\sc Verb Phrase}] {\sc Pron.Poss.3rd}$_2$ {\sc Object} {\sc Prep} {\sc Noun.DEF}. $\not\rightarrow$ {\sc Occupation.DEF.POSS}$_1$ {\sc Object} [{\sc Verb Phrase.PASSIVE}] {\sc Prep} {\sc Noun.DEF}.
\end{itemize}
}
%\ldots 

We will primarily be interested in the rate at which state-of-the-art NLI models (wrongly) predict examples of the form in Example (5) to be cases of entailment, and how this depends on whether the possessive pronoun {\sc Pron.Poss} is masculine or feminine. To generate examples of this form, we translate one prototype example and then identify the variables in the output example. We also make sure to check that there are no morpho-syntactic dependencies, e.g., agreement, between these variables. We then generate all possible examples and have native speakers manually verify the correctness of samples of the generated examples. 

\paragraph{Machine Translation} For machine translation, we are interested in the way that gender assumptions play a role in the resolution of the gendered possessive pronoun in the source language. As an example, when translating the phrase \textit{The doctor put the book on her table}, an English-Danish translation system would likely generate one of the following two options, a reflexive reading and an anti-reflexive one:

{\small
\begin{itemize}
\item[(7)] Lægen lagde bogen på \textit{sit} bord 
\item[] doctor.{\sc def} put book.{\sc def} on {\sc Pron.Poss.Refl.3rd} table

\item[(8)] Lægen lagde bogen på \textit{hendes} bord 
\item[] doctor.{\sc def} put book.{\sc def} on {\sc Pron.Poss.3rd} table
\end{itemize}
}

%While both translations are possible out of context, models would exhibit a clear gender bias if they prefer one reading over the other in the context of female entities, and the other reading in the context of male entities. 

While ABC focuses on translating {\em from English}, it holds that similarly, if we translate the Danish sentence \textit{mekanikeren har brug for sine.\textsc{refl} værktøjer til at arbejde}, which uses a gender-neutral reflexive possessive pronoun \textit{sine}, into English, the model will have to choose between two possible, correct translations:

{\small
\begin{itemize}
\item[(9)] The mechanic needs \textit{his} tools to work

\item[(10)] The mechanic needs \textit{her} tools to work
\end{itemize}
}

The machine translation section of the ABC dataset consists of translations from English sentences with gendered possessive pronouns into one of the four target languages (Danish, Russian, Swedish, and Chinese). 
For a single occupation on the list, this would correspond to two English sentences (masculine and feminine possessive pronoun) per template. We quantify to what extent models translate English source sentences with possessive masculine or feminine pronouns into target sentences with reflexive pronouns.\footnote{In the context of examples such as Example (9) and (10), using an anti-reflexive pronoun in the target translation may seem more like a hallucination than violating grammatical constraints, and we acknowledge that in machine translation, as well as in language modeling, the difference concerning existing gender bias challenge datasets is less pronounced than with NLI and coreference resolution. Nevertheless, note that the model not only hallucinates a gender attribution, but also co-referentiality, making it relatively simple to construct semantically impossible examples, e.g., {\em The mechanic needs his tools, but not his own tools}. Furthermore, introducing a new referent without evidence also violates pragmatic economy principles \cite{Grosz:ea:95,Gardent:Webber:01}. Google Translate incorrectly translates into a sentence with two reflexive pronouns (violating the semantic principle of bivalence).}

\paragraph{Coreference Resolution} 
For coreference resolution, we generate variants of our templates in the four target languages with each of the gendered anti-reflexives and the reflexive pronoun. That is , for a sentence such as: 

{\small
\begin{itemize}
    \item[(11)] {The firefighter placed \textit{her/his} shoes in the closet}
\end{itemize} 
}

we generate the following examples for Danish:
{\small
\begin{itemize}
    \item[(12)] 
Brandmanden placerede \textit{hendes} sko i skabet ({\sc Fem})
    
    \item[(13)] 
Brandmanden placerede \textit{hans} sko i skabet ({\sc Masc})
    \item[(14)] 
Brandmanden placerede \textit{sine} sko i skabet ({\sc Refl})
\end{itemize}
}

In Examples (12) and (13), the use of anti-reflexive pronouns {\em hans} or the femine anti-reflexive {\em hendes} means the shoes placed in the closet belong to someone other than the firefighter. In our coreference resolution experiments, we are thus interested in how often models wrongly link the anti-reflexive pronouns (\textit{hans/hendes}) to the occupation. Such predictions violate grammatical constraints and are clear examples of gender assumptions overwriting morpho-syntactic evidence.

\paragraph{Language Modelling} For language modeling, we are interested in how likely the models are to predict gendered anti-reflexive possessive pronoun when the original sentence contains a reflexive pronoun. In: 

{\small
\begin{itemize}
    \item[(15)] 
Brandmanden placerede \textit{sine} sko i skabet ({\sc Refl})
\end{itemize}
}
we compute the sentence perplexity replacing the reflexive pronoun {\em sine} with a feminine anti-reflexive (\textit{hendes}) or masculine (\textit{hans}) anti-reflexive pronoun. A difference in perplexity reveals a gender bias, and if the model prefers an anti-reflexive reading, this possibly leads to a grammatically incorrect sentence.\footnote{See also the footnote above on whether our machine translation examples diagnose model 'hallucinations' or unambiguous prediction errors.}

\section{Experiments}
In this work, we are focused on highlighting a linguistic phenomenon that is useful for diagnosing gender bias, therefore we do not focus on an extensive comparison of model architectures; further work would be required to examine more models. We are interested in the gender associations that \textit{existing} models make. Because of this, we take off-the-shelf translation models and language models. As there were not any state-of-the-art models already pre-trained for coreference in the languages of interest, we train a state-of-the-art architecture for coreference resolution on languages where we could obtain data. To be able to evaluate NLI models on the target languages, we fine-tune a pre-trained model for this task. 

As previously found in \cite{rudinger2018gender}, gender biases in models tended to correlate with labor statistics of the percentage of female in each occupation according to Bureau of Labor Statistics 2015 \footnote{\url{http://www.bls.gov/cps/cpsaat11.htm}} released with \newcite{caliskan2017semantics}. We correlate our findings with these statistics as well as national statistics.

\paragraph{NLI} NLI is originally a three-way classification task. Given two sentences; a premise and a hypothesis, the system classifies the relation between them as \texttt{entailment, contradiction}, or \texttt{neutral}. Since ABC is only intended for diagnosing gender bias in off-the-shelf models, and not for training models, we only consider the \texttt{entailment} relation. If the premise contains a reflexive pronoun, the true class is \texttt{entailment}, and if the premise contains a masculine or feminine pronoun it is not \texttt{entailment}. 

XNLI \cite{conneau-etal-2018-xnli} is a manual translation of the English NLI data into 15 languages. Chinese and Russian are among them and we benchmark the model on the XNLI test set. \newcite{Singh:ea:19} extend the XNLI train set to a wider set of languages, including Danish and Swedish but there is not test set for benchmarking. 
We use cross-lingual language model pre-training (XLI) \cite{lample2019cross}, i.e., we fine-tune on English NLI training data. For Chinese and Russian, we use a publicly available implementation\footnote{\url{https://github.com/facebookresearch/XLM}} of the XLM-15 model \cite{lample2019cross} and fine-tune it using a batch size of 4 and a learning rate of 0.000005 for 35 epochs, which led to the best performance on the XNLI development set. For Danish and Swedish, we use the XLM-100 model, which we fine-tune for 28 epochs. 

\paragraph{Machine Translation}
For machine translation, we evaluate models for English$\rightarrow\{$Danish, Russian, Swedish, Chinese$\}$ \textit{to assess how often they predict the non-gendered reflexive possessive pronouns when the source possessive pronoun is masculine versus feminine}. For all languages, we report the performance of Google Translate. Additionally, for the languages where an off-the-shelf, near-state-of-the-art system was publicly available, we also report performance. For Chinese, we use the pre-trained models provided by \newcite{uedin-nmt:2017} \footnote{\url{https://github.com/EdinburghNLP/nematus}} (E-WMT). For Russian, we use the winner system of WMT19 \cite{ng2019facebook}, which is provided as part of the Fairseq toolkit (F-WMT).\footnote{\url{https://github.com/pytorch/fairseq/tree/master/examples/wmt19}}  

\paragraph{Coreference Resolution} For coreference resolution, we are interested in whether the system violates grammatical rules by placing an anti-reflexive possesive pronoun in a cluster. We train coreference resolution models for Chinese and Russian using the model and code of \newcite{joshi2019spanbert}. For Chinese, we use the Chinese version of Ontonotes as our training data, which is made up of about 1800 documents for training. For Russian, we use the RuCor corpus \cite{ju2014ru}, which is small, containing only 181 documents total, but has been used to train coreference models for Russian before \cite{ju2014ru,sysoev2017coreference}. The task consists of predicting the spans that make up a coreference cluster. We train the model using the hyperparameters specified in the source code \footnote{\url{https://github.com/mandarjoshi90/coref/blob/master/experiments.conf}}. We use a maximum segment length of 128. See Appendix \S{B} for statistics of the coreference resolution datasets used for training. While we do not have coreference resolution systems we can evaluate for Danish and Swedish, we include challenge examples for these languages that can be used to detect bias in future systems for these languages.    

\paragraph{Language Modeling}
For our language modeling experiments, we use the pre-trained BERT masked language modeling architecture \cite{devlin2018bert}. We turn pronoun prediction into a Cloze task \cite{taylor1953cloze} where the pronoun is masked and then the probabilities of each possible alternative taken to compute the sentence-level perplexity. We use Chinese BERT (for Chinese) and multilingual BERT for Russian, Danish, and Swedish.\footnote{\url{https://github.com/google-research/bert/blob/master/multilingual.md}} The overall perplexities of these models on our challenge examples, are low; again, this is because of the simple vocabulary and constructions used in the examples. We nevertheless see strong gender bias in the language models, especially for Danish and Chinese.

\begin{table*}
\centering
%\small
{
\centering
\begin{tabular}{@{}lllccc@{}}
\toprule
     \textbf{Task} & \textbf{Lang} & \textbf{System} &  \textbf{Benchmark} & \textbf{ABC} &{\bf Significance}\\
     \midrule
         \multirow{4}{*}{\sc NLI}& da & XLM-100& -- & 0.380 &$\checkmark$\\
         & ru &XLM-15& 0.736 & 0.370 &$\checkmark$\\
     & sv & XLM-100& -- & 0.362 &$\checkmark$\\
    &zh & XLM-15&0.742 & 0.330 &$\checkmark$\\
    \midrule
     \multirow{6}{*}{\sc MT} &da &Google Translate& 0.204 & 0.395&  $\checkmark$\\
     \cmidrule{2-6}
          & \multirow{2}{*}{ru} & Google Translate& 0.260 & 0.406 &$\checkmark$\\
          &&F-WMT&0.268 &0.421&$\checkmark$\\
     \cmidrule{2-6}
     & sv & Google Translate& 0.211 & 0.422 &$\checkmark$\\
     \cmidrule{2-6}
     & \multirow{2}{*}{zh} & Google Translate & 0.460 & 0.594 &$\dagger$\\
     && E-WMT & 0.360 &0.194&$\checkmark$\\
     \midrule

    \multirow{2}{*}{\sc Coref}&ru & \multirow{2}{*}{\sc e2eCoref-BERT}& 0.602 & 0.090 &$\dagger$\\
    &  zh &&0.630&0.600&$\checkmark$\\
     \midrule
      
      \multirow{4}{*}{\sc LM} & da & \multirow{4}{*}{\sc BERT} & 2.4 & 11.4 &$\checkmark$\\
          & ru &  &3.9 & 13.4 & $\checkmark$\\
     & sv & &1.2 &11.2& $\checkmark$\\
     & zh & &6.7 & 22.1 &$\checkmark$\\

     \bottomrule
 \end{tabular}
 }
\caption{\label{tab:pearsonr}{\sc Gender Bias Results}. Performance on benchmarks and ABC. We look at differences in performance between genders for each task and correlate these with labor market statistics. $\checkmark$: Pearson's $\rho$ of error $\Delta$ on sentences with feminine pronouns and \%~ of women in corresponding occupations significant ($p<0.01$); see \S5~for a discussion of the labor market statistics. $\dagger$: Systems insensitive to variation in pronouns.
}
\end{table*}

\section{Results}
Our evaluation results are found in Table~2, with results on Danish (da), Russian (ru), Swedish (sv), and Chinese (zh), and for machine translation (MT), natural language inference (NLI), coreference resolution ({\sc Coref}), and language modeling (LM).

\paragraph{NLI} For NLI, the XLM models generally over-predict entailment for anti-reflexive pronouns. The models perform well on benchmark data, e.g., 0.742 on the Chinese XNLI test set, but much worse (0.330) on our challenge examples. For Chinese and Danish, the models perform slightly better on sentences with masculine anti-reflexive pronouns, whereas they perform slightly better on sentences with feminine anti-reflexives in Russian and Swedish. For all four languages, we see significant negative correlations between relative error increase on sentences with feminine pronouns and the ratio of women in corresponding occupations; see \S5 for how a discussion of the statistics. This suggests that the very poor performance numbers on sentences with anti-reflexive pronouns is, in part, the result of gender bias. 

\paragraph{Machine translation}

 For machine translation, we also observe strong negative correlations, suggesting gender bias. In the manual analysis of the output translations, we see a very clear pattern that English masculine possessive pronouns are more likely to translate into reflexive pronouns in the target languages, than feminine possessive pronouns. For Danish, 93.7\%~of masculine pronouns were translated into reflexives, whereas only 72.9\%~ of feminine pronouns were. For Russian, the two systems were consistent in this respect and both translated 69.3\% of masculine pronouns and 18.1\% of feminine pronouns into reflexive pronouns. For Swedish, the numbers were 90.0\% and 73.1\%, respectively. For Chinese, where the reflexive pronoun is used less frequently,\footnote{The systems are trained on a combination of traditional and simplified Chinese; the latter variant does not include the reflexive pronoun.} the machine translation models only produced a few translations with reflexive pronouns (for masculine source pronouns).

These differences are not reflected in BLEU scores, and in our correlations we correlate the increase in pronoun translation errors for source sentences with feminine pronouns and the ratio of women in the corresponding occupations. In general, our models achieve high BLEU scores on our challenge examples, which are all syntactically simple and use simple, in-vocabulary words.

\paragraph{Coreference Resolution}
%we see a slight bias and poor overall performance
For coreference resolution, we observe clear performance differences between our Chinese and Russian models. This possibly reflects the fact that the Russian model was trained on a very small dataset and is less likely to generalize. For both models, we observe a clear bias towards clustering masculine anti-reflexive pronouns with their grammatical subjects, despite how this violates grammar. The Chinese model, which exhibits a strong gender bias, errs on 17\%~of sentences with masculine anti-reflexive pronouns, and on 14.6\%~of sentences with feminines anti-reflexives. For Russian, the differences are small, but note the model is trained on limited data, e.g., 140 documents. Out of around 13,000 examples, the model only predicts clusters for 475 pronouns, and 400 of those are in reflexive case. The remaining 75 are masculine (0 feminine). In other words: we see a similar tendency to Chinese, but since the overall performance is poor, and the model is in general rather insensitive to differences in pronouns, we do not include correlation results.

\paragraph{Language Modeling}

Also for language modeling, we observe consistent bias when predicting a masculine pronoun in place of a reflexive for all languages. These differences are higher for Chinese and Russian.  We are not interested in the model's ability to generate a particular pronoun, the more interesting observation is whether the perplexities for sentences containing masculine possessives are lower than for predicting feminine possessives when forcing the model to predict these in place of a reflexive. Our results show that perplexities are lower for masculine possessives in all languages with the biggest differences of 3.7 sentence perplexity for Russian.

\section{Analysis: Biased statistics?}
We used occupations from \newcite{caliskan2017semantics} in creating our template data; this database also includes U.S. occupation statistics. In our results in Table~\ref{tab:pearsonr}, however, we rely on national statistics instead, but how much of a bias would it be to rely on the original American statistics? In this section, we explain how we collected the national statistics and show how they strongly correlate with the American statistics, but also that national statistics are slightly better at detecting gender bias:% in our models:

Our Danish labor market statistics come from \citet{larsen15konsopdelt}, as well as Statistics Denmark\footnote{\url{www.dr.dk/nyheder/indland/}} %ligestilling-se-hvordan-det-gaar-i-din-branche}}
and Bevægelsesregisteret,\footnote{\url{www.esundhed.dk/home/registre/}} %bevaegelsesregisteret}}
which is a national database over authorised health staff. Some numbers (paramedic, scientist and receptionist) are based on graduation statistics. The Russian labor market statistics were mostly obtained from the Federal State Statistic Service.\footnote{\url{eng.gks.ru/}} For occupations not contained on this website we obtained the numbers from separate sources such as the Center of Fire Statistics (CFS) of International Association of Fire and Rescue Services (CTIF)\footnote{\url{www.ctif.org/}} and the Organisation for Economic Cooperation and Development's statistics website\footnote{\url{stats.oecd.org/}}. We obtain most of our Swedish labor market statistics from Statistics Sweden (SCB).\footnote{\url{www.scb.se/}} We use the most recent statistic from 2017, which considers people aged 16-64 \cite{eriksson2017}. For clerk and worker, we found labor market statistics in \newcite{Women_men_in_Sweden}. For medical jobs, we used member statistics by Swedish Medical Association (SLF) from 2016.\footnote{\url{slf.se/app/uploads/2018/04/}} 
Finally, we obtain statistics for China from \newcite{Women_men_in_China}, which is based on census data from 2000.\footnote{We did not find reliable gender statistics for all occupations for all countries, but for 51 (Denmark), 50 (Sweden), 38 (Russia), and 10 (China) occupations. 
One reason was a mismatch between how gender statistics are reported in official reports, including how jobs are grouped. We release the numbers we were able to collect and will continuously work on obtaining more statistics.}

\begin{figure}
    \centering
    \includegraphics[width=3.5in]{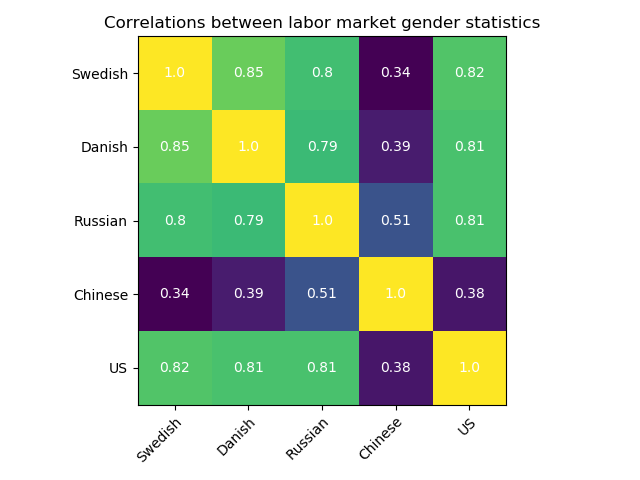}
    \caption{Correlations between collected labor statistics. Numbers $>0.7$ are significant ($p<0.01$).}
    \label{tab:international_labor_stat_correlation}
\end{figure}

While labor statistics correlate strongly across countries (Table \ref{tab:international_labor_stat_correlation}), U.S. statistics are not universal; e.g., almost all pathologist in the U.S. are women (97.5\%), whereas the percentage for Denmark is 60\%. In the U.S. and Sweden, the painter profession is very male-dominated, like mechanic and electrician (5.70\% and 8\% women, respectively), whereas in Russia, 57.0\% of painters are women. 

\paragraph{Correlation Results}
To assess the potential bias of using U.S. labor market statistics in multilingual experiments, we correlate the gender bias of models for language $l$ with labor statistics from the U.S. and the country in which $l$ is a national language, i.e., we correlate performance differences on Swedish ABC examples with both U.S. and Swedish labor statistics, Danish ABC examples with U.S. and Danish labor statistics, etc. We do so for the subset of occupations, where national gender statistics are available: \textbf{NLI} Correlations were stronger with national rather than U.S.~statistics for Danish and Swedish (-0.35 vs.~-0.28; -0.36 vs.~-0.34). 
\textbf{Machine Translation} Correlations were stronger with national rather than U.S.~statistics for Russian and Swedish (-0.31 vs.~-0.20; -0.31 vs.~-0.14). 
\textbf{Coreference Resolution} For coreference, we were able to correlate only the results for Chinese due to the fact that the coreference model for Russian only predicted clusters for sentences with male pronouns. The correlations with U.S. and Chinese labor market statistics were not significantly different because we only had statistics for 10 occupations.
\textbf{Language Modeling}
Correlations were stronger with national rather than U.S.~statistics on average, but not significantly so. 

%\begin{table}[h!]\centering
%\begin{tabular}{l|cc}
%\toprule
%Lang& National stats&  U.S. stats   \\
% \midrule
%sv & -0.12 & -0.05  \\
%ru & -0.03 &  -0.07 \\  
%zh &-0.43& -0.60 \\  
%da &-0.21 & -0.18    \\
% \bottomrule

\section{Related Work}

The ABC dataset is not first to focus on pronouns and gender bias. The UD English-Pronouns\footnote{\url{universaldependencies.org/}} \cite{Munro:20}, a manually constructed, gender-balanced benchmark of English sentences with pronouns, was motivated by the observation that the genitive pronoun \textit{hers} only occurs three times in the English Universal Dependencies \cite{nivre-etal-2016-universal}.
The gendered, ambiguous pronoun (GAP) dataset \cite{webster-etal-2018-mind} is a coreference resolution dataset of human-annotated ambiguous pronoun-name examples from English Wikipedia. \citet{prates2018assessing} constructed a translation challenge dataset of simple sentences in gender-neutral languages such as Hungarian and Yoruba and English target sentences such as {\em he/she is an engineer} to estimate gender biases in machine translation. Both these challenge datasets focus on gender hallucinations, not unambiguous errors induced by gender bias.  
Some of our examples share similarities with the English WinoGender schema \citep{rudinger2018gender}. Consider the following minimal pair of Winograd schema taken from their paper: 

{\small
\begin{itemize}
    \item[(16)] The paramedic$_1$ performed CPR on the passenger$_2$ even though {\sc Pron}$_1$ knew it was too late.
    \item[(17)] The paramedic performed CPR on the passenger$_2$ even though {\sc Pron}$_2$ was already dead.
  
\end{itemize}
}
 
In the Winograd schema, the context, i.e., the second clause, is supposed to disambiguate the pronoun on semantic grounds. In Example (16), the pronoun refers to the paramedic, because the patient is unlikely to know whether CPR is too late. In Example (17), the pronoun refers to the patient, because it is impossible to perform CPR if you are dead. Our examples, in contrast, do not disambiguate pronouns on semantic grounds and this is why we are interested in reflexive possessive pronouns: they always refer to the subject, and their anti-reflexive counterparts never do, so there is no grammatical ambiguity. The disadvantage with semantic disambiguation, we argue, is that it ultimately becomes a subjective competition of belief biases. It is generally impossible to perform CPR if you are dead, but special cases exist: 

\begin{small}
\begin{itemize}
    \item[(18)] Dr Jones$_1$ has turned into a zombie! He$_1$ performed CPR on the passenger even though he$_1$ was already dead. 
\end{itemize}
\end{small}

The ABC dataset evaluates to what extent gender bias leads to unambiguous NLP errors not based on semantic grounds. Finally, \newcite{zhao2018gender} also include English examples with reflexive pronouns that can be resolved on syntactic grounds, such as: 

{\small
\begin{itemize}
    \item[(19)] The secretary called the physician and told {\it him} about a new patient.
\end{itemize}
}

This construction, however, is less interesting than the reflexive possessive pronominal construction, since in this case, pronouns are always co-referential with the object position, regardless of the pronoun. In sum, the ABC challenge dataset is, to the best of our knowledge, the first dataset to focus on cases where gender bias leads to unambiguous errors; it is also the first multilingual, multi-task gender bias challenge dataset, and the first to focus on anti-reflexive pronouns.  

\section{Conclusion}
In this work we have introduced the Anti-reflexive Bias Challenge (ABC) dataset for multilingual, multi-task gender bias detection, the first of its kind, including four languages and four tasks: machine translation, natural language inference, coreference resolution and masked language modeling. The ABC dataset focuses on a specific linguistic phenomenon that does not occur in English but is found in languages with \textit{Type B reflexivization}: namely, anti-reflexive gendered pronouns. This phenomenon is shown to be useful for exposing unambiguous gender bias, because it quantifies to what extent gender bias leads to prediction {\em errors}, in contrast to just {\em unwarranted disambiguations} ('hallucinations'). The problem of anti-reflexive gendered pronouns has, to the best of our knowledge, not received attention before in the NLP literature, which tends to focus heavily on English \cite{Bender:Friedmann:18}. Our evaluations of state-of-the-art models across the four tasks generally reveal significant gender biases leading to false predictions. Additionally, we find that for some tasks, these associations are more in line with national labor market gender statistics than with U.S. statistics, revealing another way that anglocentric biases can prohibit the detection of gender biases in our models.

\section*{Acknowledgements}
We want to thank Heather Lent, Dustin Wright, Sheng Zhang and Daniel Hershcovich as well as the anonymous reviewers for their valuable feedback. This work was supported by Google Focused Research Award, as well as a performance contract allocated to the Alexandra Institute by the Danish Ministry of Higher Education and Science.

\bibliography{anthology,emnlp2020}
\bibliographystyle{acl_natbib}

\appendix
\clearpage
\section{Example Data}
We present example data for each task. 

\subsection{NLI}
For each occupation, template and language in our NLI benchmark, we have three premise-hypothesis pairs. The hypotheses are identical. For the premises, we have three versions, one for each pronoun (the non-gendered reflexive possessive pronoun and and the two gendered, anti-reflexive possessive pronouns). Examples with reflexives are labeled {\tt entailment}; examples with anti-reflexives are labeled {\tt neutral}. An example datapoint for this task, with the label {\tt entailment}, is shown in Table \ref{tab:example_NLI}.

\begin{table}[h!]\centering
\centering

\begin{tabular}{p{.3cm}|p{6cm}}
\toprule

\multicolumn{2}{c}{{\sc Natural language inference}}\\
\midrule
\multicolumn{2}{c}{{\sc a) Premise and b) Hypothesis}}\\
\midrule
\multirow{4}{*}{{\bf da}}& a) Ingeniøren tager \textit{sine} briller fra skrivebordet \newline b) Ingeniørens briller tages fra skrivebordet \\%\newline Label:\textit{Entailment} \\
\midrule
\multirow{4}{*}{\bf sv} & a)  Ingenj\"{o}ren tar \textit{sina} glas\"{o}gon från skrivbordet \newline b) Ingenjörens glasögon tas bort från skrivbordet.\\% \newline Label:\textit{Entailment}\\
\midrule
\multirow{3}{*}{\bf ru}& a) Inzhener snimayet \textit{svoi} ochki so stola. \newline b) Ochki inzhener snyaty so stola.\\ %\newline Label:\textit{Entailment}\\
\midrule
\multirow{4}{*}{\bf zh}& a) Gōngchéngshī cóng zhuōzi shàng zhāi xià \textit{zìjǐ} de yǎnjìng \newline b)  Gōngchéngshī de yǎnjìng yǐ bèi cóng zhuōzi shàng yí kāi.\\ %\newline Label:\textit{Entailment}\\
\midrule
\multicolumn{2}{c}{{\sc English translation}}\\
\midrule
\multirow{2}{*}{\bf en}& a) The engineer takes \textit{his/her} glasses from the desk \newline b)  The engineer's glasses are taken from the desk\\
\bottomrule
\end{tabular}

\caption{\label{tab:example_NLI}Example data for NLI. For NLI, we only generate entailments and neutral statements. The English translation is shown for reference only.}
\end{table}

\subsection{Machine Translation}
For machine translation, we have 4560 pairs of source sentences with masculine (\textit{his}) and feminine possessive pronouns (\textit{her}), respectively. We translate these into the target languages using off-the-shelf models and assess the tendency of these models to predict reflexive possessive pronouns in the target languages, instead of anti-reflexive possessive pronouns. An example of the data can be found in Table \ref{tab:example_mt}. 

\begin{table}[h!]\centering
\centering

\begin{tabular}{p{.3cm}|p{6cm}}
\toprule

\multicolumn{2}{c}{{\sc Machine translation}}\\
\midrule
\multicolumn{2}{c}{{\sc Source sentence}}\\
\midrule
{\bf en}&The engineer takes \textit{his/her} glasses from the desk\\
\midrule
\multicolumn{2}{c}{{\sc Translations}}\\
\midrule
\multirow{2}{*}{\bf da} & Ingeniøren tager \textit{sine} briller fra skrivebordet\\
\midrule
\multirow{2}{*}{\bf sv} & Ingenj\"{o}ren tar \textit{sina} glas\"{o}gon från skrivbordet\\
\midrule
\multirow{1}{*}{\bf ru}& Inzhener snimayet \textit{svoi} ochki so stola.\\
\midrule
\multirow{2}{*}{\bf zh}& Gōngchéngshī cóng zhuōzi shàng zhāi xià \textit{zìjǐ} de yǎnjìng\\

\bottomrule
\end{tabular}

\caption{\label{tab:example_mt}Example data for machine translation. }
\end{table}

\subsection{Coreference Resolution}

For coreference resolution, we are interested in whether the model is more likely to cluster a masculine possessive pronoun with the subject of the sentence than a feminine pronoun, even when this reading violates grammatical constraints. In Table \ref{tab:example_coref}, we list examples of how the task data would look. In brackets, we have mentions of an entity that can be clustered together by the system as belonging to the same coreference chain.
\begin{table}[h!]\centering
\centering

\begin{tabular}{p{.3cm}|p{6cm}}
\toprule

\multicolumn{2}{c}{{\sc Coreference resolution}}\\
\midrule

\multirow{2}{*}{\bf da}& [Ingeniøren] tager \textit{[sine/hans/hendes]} briller fra skrivebordet \\
\midrule\multirow{2}{*}{\bf sv}&[Ingenj\"{o}ren] tar \textit{[sina/hans/hennes]} glas\"{o}gon från skrivbordet\\
\midrule\multirow{2}{*}{\bf ru}& [Inzhener] snimayet \textit{[svoi/yego/yeye]} ochki so stola.\\
\midrule\multirow{2}{*}{\bf zh}& [Gōngchéngshī] cóng zhuōzi shàng zhāi xià \textit{[zìjǐ/tā/tā]} de yǎnjìng \\
\midrule
\multicolumn{2}{c}{{\sc English translation}}\\
\midrule
\multirow{2}{*}{\bf en}& [The engineer] takes \textit{[his/her]} glasses from the desk \\
\bottomrule

\end{tabular}

\caption{\label{tab:example_coref} Example data for coreference resolution. In brackets, we have the mentions that the system could cluster as coreferent. We include an English translation only for reference.}
\end{table}

\subsection{Language Modeling}

For language modeling, we take a sentence containing a reflexive pronoun and swap the reflexive for the possessive masculine and feminine anti-reflexives; we then compute the perplexities of the original and perturbed sentences. Example of how this is framed can be found in Table \ref{tab:example_lm}.

\begin{table}
\begin{tabular}{p{.3cm}|p{6cm}}
\toprule

\multicolumn{2}{c}{{\sc Language modeling}}\\
\midrule 
\multirow{6}{*}{\bf da}& \textbf{Truth}: Ingeniøren tager \textit{sine} briller fra skrivebordet \newline \textbf{Prediction(Fem)}: Ingeniøren tager \textit{hendes} briller fra skrivebordet \newline  \textbf{Prediction(Masc)}: Ingeniøren tager \textit{hans} briller fra skrivebordet \\
\midrule 
\multirow{6}{*}{\bf sv}& \textbf{Truth}: ingenj\"{o}ren tar \textit{sina} glas\"{o}gon från skrivbordet \newline \textbf{Prediction(Fem)}: ingenj\"{o}ren tar \textit{hennes} glas\"{o}gon från skrivbordet \newline  \textbf{Prediction(Masc)}: Ingenj\"{o}ren tar \textit{hans} glas\"{o}gon från skrivbordet \\
\midrule 
\multirow{6}{*}{\bf ru}& \textbf{Truth}: Inzhener snimayet \textit{svoi} ochki so stola. \newline\textbf{ Prediction(Fem)}: Inzhener snimayet \textit{yeye} ochki so stola. \newline  \textbf{Prediction(Masc)}: Inzhener snimayet \textit{yego} ochki so stola. \\
\midrule 
\multirow{7}{*}{\bf zh}& \textbf{Truth}: Gōngchéngshī cóng zhuōzi shàng zhāi xià \textit{zìjǐ} de yǎnjìng \newline \textbf{Prediction(Fem)}: Gōngchéngshī cóng zhuōzi shàng zhāi xià \textit{tā} de yǎnjìng \newline  \textbf{Prediction(Masc)}: Gōngchéngshī cóng zhuōzi shàng zhāi xià \textit{tā} de yǎnjìng \\
\bottomrule
\end{tabular}

\caption{\label{tab:example_lm}Example data for the language modeling task }
\end{table}

\section{Coreference Dataset Statistics}
In table \ref{tab:coref_stats} we show the number of documents used to train each system. For Chinese, the data is available with predefined train, development and test sets. For Russian, however, this is not specified, therefore we split the data 80-20-20.
\begin{table}[h]\centering
\footnotesize
\begin{tabular}{l|rrr}
\toprule
 Lang&  Training&  Dev & Test  \\
 \midrule

zh & 1810 & 252 &   218   \\
ru & 144 & 18 & 18    \\
 
 \bottomrule

\end{tabular}
 \caption{\label{tab:coref_stats}Statistics for the coreference data used for training.  }
\end{table}

\end{document}